\documentclass[letter, 10 pt, conference]{ieeeconf}  % Comment this line out if you need a4paper
\usepackage{scalerel}
\usepackage{hyperref} %<--- Load after everything else
\usepackage{epsfig}
\usepackage{epstopdf}
\usepackage{amsmath}
\usepackage{amssymb}
\usepackage{bm}
\usepackage{graphicx} 
\usepackage{booktabs}

\usepackage[font=footnotesize,skip=4pt]{subcaption}
\usepackage[font=footnotesize,skip=2pt]{caption}

\usepackage{algorithm}
\usepackage{lipsum}
\usepackage{graphicx}
\IEEEoverridecommandlockouts                              % This command is only needed if 
\usepackage{hyperref}
\usepackage{cleveref}
\usepackage{scalerel}
\usepackage{tikz}
\usepackage{algpseudocode}
\algnewcommand{\LineComment}[1]{\State /* \textit{#1} \hfill */}

\usetikzlibrary{svg.path}

\definecolor{orcidlogocol}{HTML}{A6CE39}
\tikzset{
	orcidlogo/.pic={
		\fill[orcidlogocol] svg{M256,128c0,70.7-57.3,128-128,128C57.3,256,0,198.7,0,128C0,57.3,57.3,0,128,0C198.7,0,256,57.3,256,128z};
		\fill[white] svg{M86.3,186.2H70.9V79.1h15.4v48.4V186.2z}
		svg{M108.9,79.1h41.6c39.6,0,57,28.3,57,53.6c0,27.5-21.5,53.6-56.8,53.6h-41.8V79.1z M124.3,172.4h24.5c34.9,0,42.9-26.5,42.9-39.7c0-21.5-13.7-39.7-43.7-39.7h-23.7V172.4z}
		svg{M88.7,56.8c0,5.5-4.5,10.1-10.1,10.1c-5.6,0-10.1-4.6-10.1-10.1c0-5.6,4.5-10.1,10.1-10.1C84.2,46.7,88.7,51.3,88.7,56.8z};
	}
}

\DeclareMathOperator{\Img}{\mathbf{I}}

\newtheorem{definition}{Definition}

\newcommand\orcidicon[1]{\href{https://orcid.org/#1}{\mbox{\scalerel*{
				\begin{tikzpicture}[yscale=-1,transform shape]
				\pic{orcidlogo};
				\end{tikzpicture}
			}{|}}}}

\title{\large \textbf{NUSense: Robust Soft Optical Tactile Sensor
}}

\author{\authorblockN{Madina Yergibay*$^{\orcidicon{0000-0002-6214-9607}}$, Tleukhan Mussin*$^{\orcidicon{0009-0007-8259-7849}}$, Saltanat Seitzhan$^{\orcidicon{0009-0002-0620-9617}}$, Daryn Kenzhebek$^{\orcidicon{0009-0000-5565-7200}}$, Zhanat Kappassov$^{\orcidicon{0000-0003-3262-3993}}$, \\ Harold Soh$^{\orcidicon{0000-0002-3278-0035}}$,Tasbolat Taunyazov$^{\orcidicon{0000-0002-0782-5553}}$}%
\thanks{This work was funded by MSHE Kazakhstan Grant number AP23485994,  by Nazarbayev University under FDCRGP no. 11022021FD2923 and 201223FD2606. \textit{Corresponding author: Zhanat Kappassov}}
\thanks{M. Yergibay, T. Mussin, D. Kenzhebek, S. Seitzhan, and Z. Kappassov are with the Robotics Department, Institute of Smart Systems and Artificial Intelligence, Nazarbayev University, Astana, Kazakhstan.}
\thanks{H.Soh and T. Taunyazov is with Dept. of Computer Science, National University of Singapore, Singapore (e-mail: tasbolat@comp.nus.edu.sg).}
\thanks{*Contributed equally.}
}
\begin{document}

\maketitle
\thispagestyle{empty}
\pagestyle{empty}
%%%%%%%%%%%%%%%%%%%%%%%%%%%%%%%%%%%%%%%%%%%%%%%%%%%%%%%%%%%%%%%%%%%%%%%%%%%%%%%%
\begin{abstract}
While most tactile sensors rely on measuring pressure, insights from continuum mechanics suggest that measuring shear strain provides critical information for tactile sensing. In this work, we introduce an optical tactile sensing principle based on shear strain detection. A silicone rubber layer, dyed with color inks, is used to quantify the shear magnitude of the sensing layer. This principle was validated using the NUSense camera-based tactile sensor. The wide-angle camera captures the elongation of the soft pad under mechanical load, a phenomenon attributed to the Poisson effect. 
The physical and optical properties of the inked pad are essential and should ideally remain stable over time. We tested the robustness of the sensor by subjecting the outermost layer to multiple load cycles using a robot arm. Additionally, we discussed potential applications of this sensor in force sensing and contact localization.

% While the most of tactile sensors depend on measuring pressure, implications derived from continuum mechanics  suggest for the measurement of shear strain. We introduce an optical tactile sensing principle based on this shear strain detection. A silicone rubber dyed with color inks is used to quantify the shear magnitude of the sensing layer. The principle is validated using a camera-based tactile sensor, NUSense.  The camera with a wide angle of view captures the elongation -- owing to the Poisson effect -- of the sensor's soft pad under a mechanical load. Physical and visual properties of the inked pad are crucial and, ideally, should not change with the time. We tested the robustness of sensor on a bench using a robot arm to expose the outermost layer to multiple cycles. We also discuss putative applications of this sensor for force sensing and  contact localization.
\end{abstract}

% \begin{abstract}
% While most tactile sensors depend on measuring pressure, implications derived from continuum mechanics suggest the measurement of shear strain. We introduce an optical tactile sensor, NUSense, based on this principle of shear strain detection. A silicone rubber dyed with color inks is used to quantify the shear magnitude of the sensing layer. A camera with a wide angle of view captures the elongation -- resulting from the Poisson effect -- of the sensor’s soft pad under mechanical load. The physical and visual properties of the inked pad are crucial and, ideally, should not change over time. We tested the robustness of the sensor on a bench using a robot arm to expose the outermost layer to multiple cycles. Additionally, we discuss potential applications of this sensor for force sensing and contact localization.
% \end{abstract}
%%%%%%%%%%%%%%%%%%%%%%%%%%%%%%%%%%%%%%%%%%%%%%%%%%%%%%%%%%%%%%%%%%%%%%%%%%%%%%%%
\section{Introduction}
\label{sec:intro}
Robust tactile sensing is generally acknowledged to be important for effective robotic manipulation. Modern robots can be equipped with soft pads that deform upon contact with objects and this deformation, which encodes the interaction properties between the robot and its environment, is a key phenomenon driving research in tactile sensing. Measuring and understanding these deformations can enable robots to handle objects dexterously and safely.
\begin{figure}[t]
\centering
\includegraphics[width=0.48\textwidth]{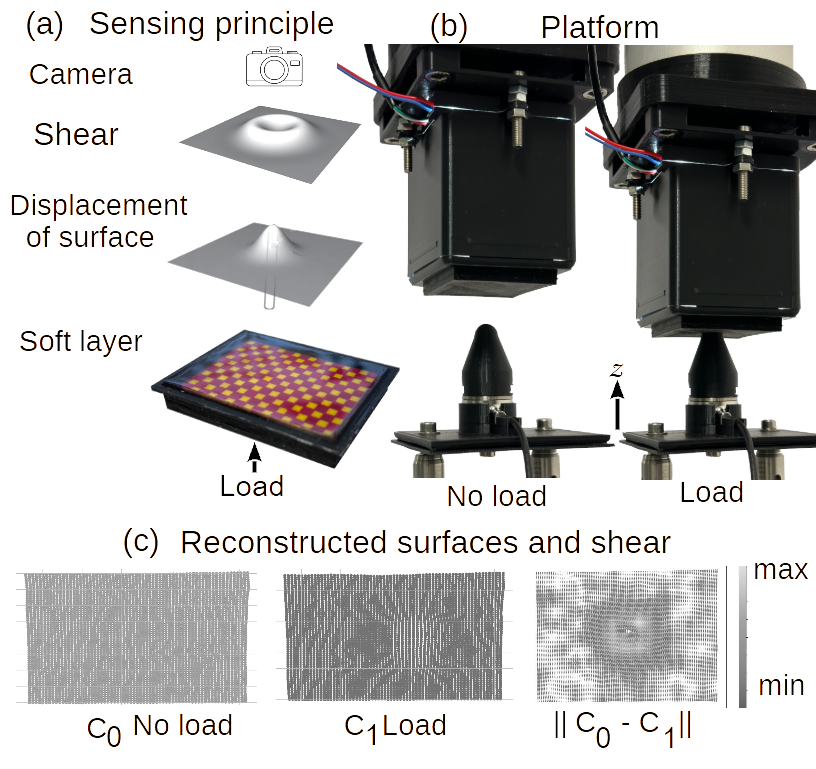}
\caption{NUSense tactile sensor. a) Sensing principle: surface displacement caused by a small axi-symmetric punch (load); the soft layer is made of silicone rubber that is dyed with color inks. The resulting pattern is projected onto camera's pixels. b) A robot arm with the sensor and the indenter for the point load. c) Surface construction using B-spline patches from camera images during resting state and under mechanical load. The region of shear deformation creates an annulus. Shear is estimated as the distance between the two surfaces represented by B-spline control nets $C_0$ and $C_1$. }
\label{fig:concept}
\end{figure}
% Successful object handling depend on the properties of contacts and within robotics, contact sensing techniques have long been recognized to be crucial for dexterous manipulation.
% However, fine and yet robust measurements of contact properties through surface deformations remains challenging. 

% Since the 1970s, there have been  attempts to create sensitive artificial tactile sensors: from a simple array of piezo-resistive load cells~\cite{belgrade_hand_skin_1977} to  complex optical sensors~\cite{optical_sensor_servo_Lepora_2024}. 
% One approach to contact detection is to imitate the features of the soft pads on human fingertips. 

However, accurate measurements of contact properties through surface deformations  remain a challenging problem. 
%A prominent approach to improving tactile sensing is xto mimic the soft pads of human fingertips. 
Since the 1970s, various sensitive artificial tactile sensors have been developed, ranging from simple arrays of piezo-resistive load cells~\cite{belgrade_hand_skin_1977} to more complex optical sensors~\cite{optical_sensor_servo_Lepora_2024}. Many existing tactile sensors measure pressure. Nevertheless, research in continuum mechanics~\cite{modeling_fem_Ricker} and human tactile perception~\cite{skin_stretch_Kikuuwe_2005}  suggest the importance of capturing shear strain rather than normal strain --- shear strain provides important information about surface interactions. Designing tactile sensors that robustly capture shear strain over multiple loading cycles is difficult. Existing sensors, such as optical tactile sensors with embedded markers, degrade over time due to wear on the outer layers, making them prone to noise.

To address these limitations, we introduce an optical tactile sensing principle based on shear strain detection (Figure~\ref{fig:concept}a). Building on our previous work~\cite{baimukashev2020shear}, we present a sensor that uses silicone rubber dyed with color inks to quantify shear deformation. The outermost layer, made of thick black silicone rubber, protects the sensor from physical damage and ambient light interference. Our image-processing algorithm estimates shear magnitude by analyzing the color pattern on the sensor's surface. This work builds on the shear sensing hypothesis established in our previous research~\cite{baimukashev2020shear} and uses the same color palette.

The primary contributions of this paper are: 1) a tactile sensing principle based on shear strain, 2) a robust sensor design and fabrication process, and 3) a model-based algorithm to extract tactile data from camera images. The remainder of this paper is organized as follows: Section~\ref{sec:related_works} provides an overview of optical tactile sensing. In Section~\ref{sec:prince}, we describe our sensing principle in detail. The design and fabrication process of the sensor is outlined in Section~\ref{sec:assembly}. Section~\ref{sec:data_proc} discusses the image-processing pipeline. In Section~\ref{sec:characterization}, we present the sensor’s properties and experimental results on force measurements and contact localization. Finally, we conclude the paper in Section~\ref{sec:concl}.

\section{Related Works}
\label{sec:related_works}
As robots become more integrated into human environments, it becomes increasingly important for them to safely and reliably manipulate everyday objects. Tactile sensors are considered essential for enabling dexterous object manipulation and thus, it is important to develop a general-purpose tactile sensor capable of meeting the demands of diverse manipulation tasks. 

Optical or vision-based tactile sensors (VBTS), particularly those with soft pads, have emerged as promising candidates due to their versatility and effectiveness in various applications. These sensors have gained significant attention over the past decades, with researchers developing various hardware and software prototypes that enable robots to acquire direct feedback from contact. Unlike conventional piezoresistive, piezoelectric, and capacitive sensors, VBTS provide higher spatial resolution, remain unaffected by electromagnetic interference, and offer expanded capabilities beyond force perception, such as texture recognition, shape detection, 3D reconstruction, and slip detection, by integrating multiple modalities within a single sensor~\cite{shimonomura2019tactile}. 

Leading advancements in the field include sensors such as GelSight ~\cite{yuan2017gelsight}, GelForce ~\cite{sato2009finger}, TacTip ~\cite{ward2018tactip}, FingerVision ~\cite{zhang2018fingervision}, and Digit ~\cite{lambeta2020digit}. %Due to the availability of high-resolution RGB cameras at low cost, coupled with the ability to incorporate various functionalities into one sensor, VBTS hold significant potential for providing tactile feedback in robotic applications.
The fundamental components of most VBTS designs include a soft sensing skin for contact, an illumination source (typically LEDs), a camera, and a 3D-printed casing to house the system ~\cite{zhang2022hardware}.

The soft skin is generally made of commercially available silicone materials, and depending on the application, it may incorporate markers, patterns, or a protective layer. In force perception and slip detection scenarios, 2D or 3D markers are often employed, with 3D-shaped or multi-colored markers enabling the measurement of shear forces ~\cite{shimonomura2019tactile}. For applications like texture classification and shape reconstruction, markerless skins are preferred to avoid visual obstruction of tactile images. However, our approach, inspired by ~\cite{baimukashev2020shear}, employs a 2D grid pattern that maintains the integrity of texture and shape data while simultaneously serving as control points for the perception of normal, shear, and torsional forces, without compromising spatial resolution. 

Lighting plays a crucial role in transducing tactile information from the skin surface to the camera. Uniform light distribution is essential for maintaining consistent sensor performance. In applications requiring depth estimation for 3D reconstruction, some designs utilize monocular photometric sensing, while others employ RGB LEDs to enhance contrast and provide rich tactile data from the sensor surface. However, most sensors rely on uniform white light in combination with markers, patterns, or colored silicone skins.%~\cite{zhang2022hardware} 
%~\cite{li2024vision}.
In certain cases, light refraction is leveraged as the primary transduction mechanism, where changes in ray angles upon contact are analyzed for segmenting contact positions and areas ~\cite{shimonomura2016robotic}.

Vision-based tactile sensors (VBTS) typically utilize compact RGB cameras to maintain a low-cost and compact form factor. Some systems have incorporated binocular stereo cameras, while others employ depth cameras. Recent developments have also explored the use of event-based neuromorphic cameras in tactile sensing~\cite{kumagai2019event, muthusamy2020neuromorphic, rigi2018novel, ward2020neurotac}. 

Advances in computer vision and image processing have significantly improved the processing of image data in VBTS. Common data processing tasks include contact segmentation, 3D reconstruction, force estimation, slip detection, and mapping and localization. For force estimation, marker displacements are tracked, and the image-derived data are processed either using traditional mathematical modeling or fed into neural networks and deep learning algorithms~\cite{li2024vision}. While mathematical modeling is more data-efficient and requires less computational power, deep learning approaches offer greater flexibility, provided a sufficiently large dataset is available.

Despite progress with low-cost RGB camera-based sensors, challenges persist in integrating multiple functions into a single VBTS. Future sensors should support diverse applications with minimal hardware changes. Additionally, more robust markers are needed to measure forces reliably without obstructing spatial resolution or distorting contact shape and texture.

\section{Shear Strain Sensing Principle}
\label{sec:prince}

Despite the significant progress on vision-based tactile sensors, it is important to note that touch inherently involves contact mechanics, which is fundamentally different from optics. As the late V. Hayward aptly noted: ``\emph{Haptics and tactile sensing is thus the province of mechanics}'' ~\cite{Hayward_said_2015}. 

%%%%inspiration from human skin
\subsection{Contact mechanics of stresses and strains in human glabrous skin.}

In the human somatosensory system, mechanoreceptors report tissue displacement to the brain. These signals are mediated by the contact mechanics of soft tissues subjected to mechanical loads. The relationship between these signals and skin deformation is a complex issue, and a full discourse is beyond the scope of this paper. In brief, Figure~\ref{fig:biomimetics}a illustrates the simplified structure of human glabrous skin under normal displacement. It is believed that the ridges in the epidermis play a role in detecting the amplitude of shear strain, allowing humans to perceive sharp edges. %~\cite{edge_haptics_2024}. 
The epidermis and dermis are interconnected via these ridges. Normal displacement of the epidermis induces tangential displacement in the dermis due to shear strain, $\gamma$.

The epidermis surface displacement results from mechanical load applied to the skin~\cite{platkiewicz_haptic_2016}. Assuming the skin as an incompressible homogeneous body, the concentrated force $F\cdot \delta(x,y) \in \Re^3$ applied to the position $(0, 0)$ causes displacement $\boldsymbol{u}(x, y, z) \in \Re^3$ at position $(x, y)$ with depth $z$. 
The mechanical response of the soft tissue, denoted as $g(x,y)$, is typically modeled as a low-pass filter, acting as a `spread function' (see surface displacement in Figure~\ref{fig:concept}a). In other words, the detection of a contact on soft skin begins by blurring it using a Gaussian filter with width $\varepsilon$, $\phi_{\varepsilon}(x,y)$.
%Given this concentrated force, the mechanical response under of the soft tissue is assumed to be $g(x,y)$. Owing to the elastic mechanics of the tissue, this response is usually assumed to have the form of a low-pass filter that acts as `spread function' (see Displacement of surface in Figure~\ref{fig:concept}a). In other words, the detection of a contact on a soft skin begins by blurring it using a Gaussian of width $\varepsilon$, $\phi_{\varepsilon}(x,y)$.%$~\cite{Pezzementi_pointspread_2010}. 

The stress distribution in the skin can be expressed as the convolution   $\sigma(x,y) = (f\ast g)(x,y)$, where $f(x,y)$ is an arbitrary pressure distribution on the skin~\cite{spatial_response_elastic_layer_1997}. For a small point load, the stress distribution in the skin induces shear deformation, forming an annular region~(see Figure~\ref{fig:concept}a)~\cite{platkiewicz_haptic_2016}. Such shear deformation can be explained using a system that produces shearing stress in two intersecting lines across which it acts. Figure~\ref{fig:biomimetics}b illustrates this planar example with a soft silicone rubber subject to purely normal displacement.  %of a part of its surface. 
The shearing stress consists of equal tangential stresses along two lines $AB$ and $A'B'$. These tangential stresses produce tangential displacements known as shear strain. By Hooke's law, the shear field at $z=0$, $\gamma(x)$, can be expressed as:
%This result can be seen by locating a narrow contact at the origin, x=0 and expressing the shear field, γ(x, z), as the convolution of the surface pressure, p(x), by a Gaussian of width ε, ϕε(x), differentiated along the spatial variable, x, scaled by depth, z and by the inverse of the elastic modulus of the medium, E
\begin{equation}
    \gamma(x) \simeq \frac{3}{E}\frac{d}{dx}(\sigma \ast \phi_{\varepsilon})(x),
    \label{eq:gamma}
\end{equation}
where $E$ is the Young's modulus of the medium.

%%%% suggestions to our sensor design
\subsection{Implications for soft pad design.}
The mechanistic model of skin described above is based on continuum mechanics theory, which is applicable to incompressible materials such as rubber. Our sensor's soft pad consists of multiple layers of silicone rubber with varying properties. As described in the next section, this pad is placed over a camera with a backlight. To enhance the image response due to shear strain, one layer of silicone rubber is dyed with colored inks.

\begin{figure}[t]
\centering
\includegraphics[width=0.4\textwidth]{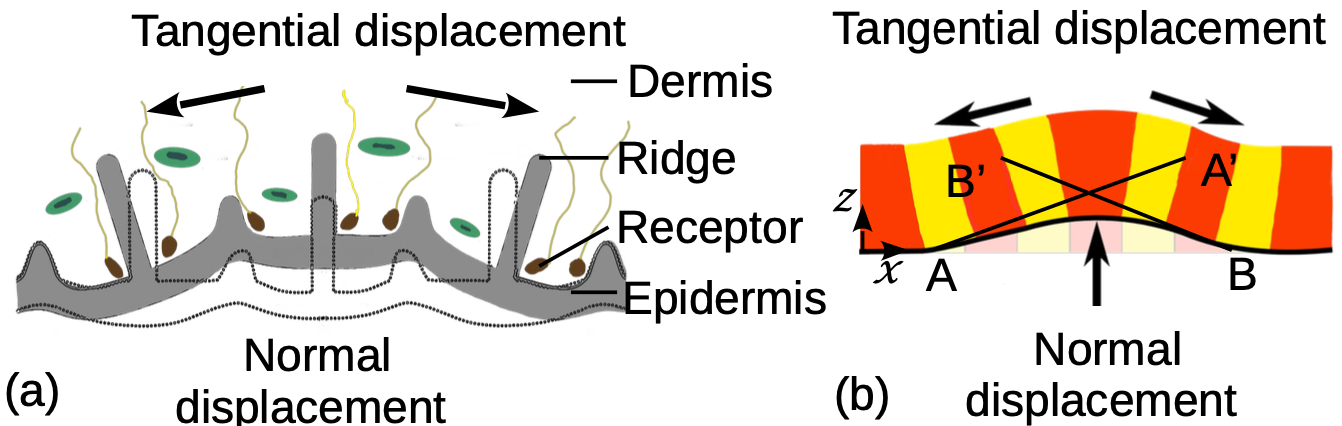}
\caption{Bio-mimetic sensing principle: a) Human glabrous skin. The epidermis has ridges projecting into the dermis in the human glabrous skin. b) The structure and mechanics of the artificial sensing skin.}
\label{fig:biomimetics}
\end{figure}

Figure~\ref{fig:biomimetics}b shows  the sensing layer made of the alternating yellow and red blobs. They shift due the tangential displacement of the soft layer subject to normal displacement. The borders between  neighboring blobs act as ridges in the human skin, as shown in Figure~\ref{fig:biomimetics}a. The circumferential area around the point of contact undergoes tensile stress and elongation, which is captured by the camera positioned beneath the colored layer.

The elongation is quantified using unique geometrical features of this pattern. These features are attached to the pattern, similar  to how ridges are attached to receptors in human skin (Figure~\ref{fig:biomimetics}a). 
This tactile sensing principle, based on shear strain detection, offers robustness to varying contact conditions, an improvement over our previous techniques~\cite{baimukashev2020shear, Mukashev_photoelastic_2022, Kappassov_rgb_2019}.
%Such tactile sensing principle based on shear strain detection is applied with benefit of invariance to contact conditions, which our techniques presented in~\cite{baimukashev2020shear, Mukashev_photoelastic_2022, Kappassov_rgb_2019} did not afford.

\section{Sensor Fabrication}
\label{sec:assembly}
%In this section, we describe the essential requirements for the sensor, its main components, and the fabrication of the soft sensing pad.
%
It is crucial to capture tactile information with clarity and sufficient spatial resolution. To achieve this, the sensor should incorporate a camera with adequate resolution, and the color grid must provide enough features to detect even subtle touches. Previous evaluations suggest that the sensor's outermost layer must be robust yet sensitive. Specifically, the outer layer should not be too thick, as this would reduce the sensor’s sensitivity~\cite{spatial_response_elastic_layer_1997}. %The difference in stress distribution $\sigma(x,y)$ for 0.6 mm and 0.2 mm thick layers is in order of 10~\cite{spatial_response_elastic_layer_1997}. textwidth

\subsection{Materials}
We fabricated the sensor using off-the-shelf components. To record the shift of color grid, a webcamera (Zerodis, HBVCAM-3M2111WA V22) acquired the images. The grid was illuminated with diffuse white 
light from surface-mount LEDs (3528) on a printed circuit board and driven
with 100 mA current each. The light was guided from four sides through glass (Plexiglass, 3 mm thick). The soft pad of the sensor was placed onto the rigid transparent glass to fix the color grid.

%The camera mount, base, and lid were 3D printed using Polylactic acid (PLA) filament.  The outermost layer was molded with silicone rubber (Smooth-On, Sorta-Clear 18). The sensing layer, which is embedded with the colored pattern, was painstakingly cast. Next layer is filled with soft clear gel (Techsil, RTV27905). Finally, in order to fasten the soft pad to glass, a thin layer of Polydimethylsiloxane (PDMS) was coated onto the latter layer. 

The camera mount, base, and lid were 3D printed using Polylactic Acid (PLA) filament. The outermost layer was molded using silicone rubber (Smooth-On, Sorta-Clear 18). The sensing layer, embedded with the colored pattern, was meticulously cast. The next layer was filled with soft clear gel (Techsil, RTV27905). Finally, a thin layer of Polydimethylsiloxane (PDMS) was used as a top layer to secure the soft pad to the glass.

\begin{figure}[h]
\centering
\includegraphics[width=1\columnwidth]{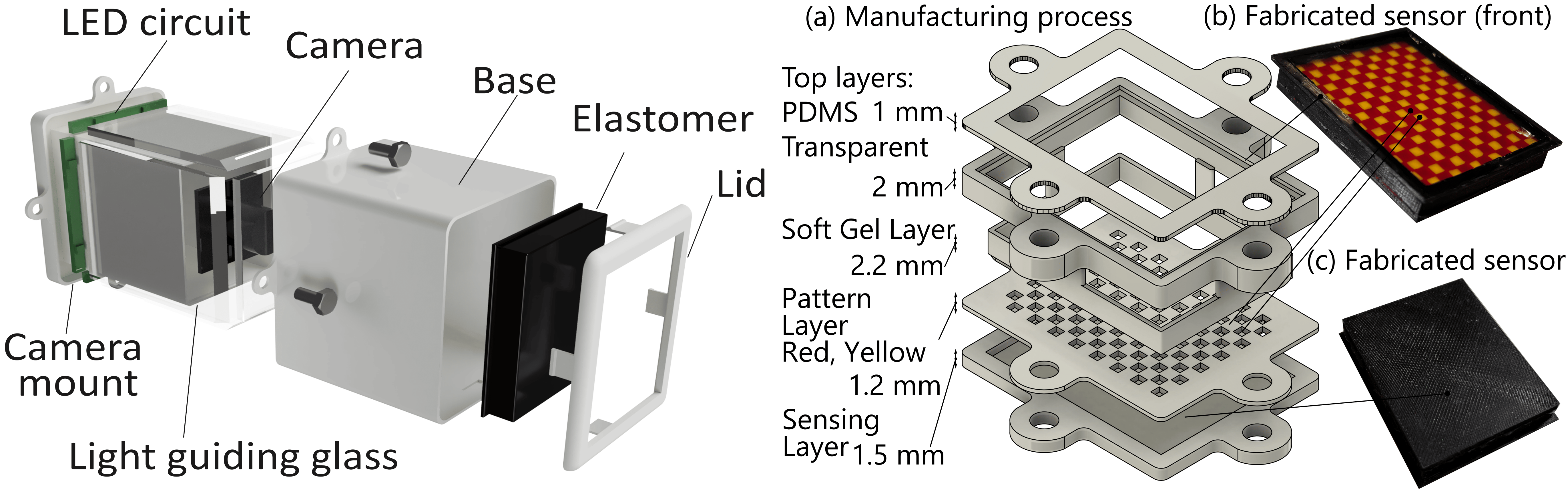}
\caption{\textbf{Left}: Assembly of NUSense tactile sensor with a camera and elastomer. \textbf{Right}: The silicone molding-based fabrication of NUSense. (a) Manufaturing process with all molding layers, (b) front and (c) back of the fabricated elastomer. }
\label{fig:assembly}
\end{figure}

Figure~\ref{fig:assembly} illustrates the hardware components of the sensor. The soft pad (Elastomer) is fastened in between the lid and the base. The LED PCB, camera, and light guiding glasses are attached to the camera mount. Since small deviations in the camera image can result in false positives, the base of the sensor should be rigidly fixed onto the camera mount. However, the sensor should also be easy to unseal securely and quickly, so we fastened it using nuts and bolts.

The sensing area is a $53\times40$ mm rectangle. The elastomer is located at the camera's focal length (55 mm). The thickness of the 3D-printed base was 1.5 mm to prevent the ambient light from affecting the internal illumination of the sensor.
\subsection{Manufacturing process}
% The fabrication of the sensing layers was guided by the aforementioned requirements, including the desire to optimize the cost of the materials.

The soft pad of our NUSense sensor is constructed layer by layer. Figure~\ref{fig:assembly}a represents the manufacturing process. Each layer of silicone is cast on top of the other. All layers are manufactured by mixing A and B components of chosen silicone rubber along with the suitable color ink for each layer.  

The process begins by casting the sensing layer using black silicon rubber. Then, the painted layer is created in two steps: 1) we placed the  silicone dyed with yellow ink into the pattern mold; 2) we removed the mold and cast the silicone dyed with red ink. Once the silicone layer with the color grid is ready, the Soft Clear Gel Layer is inserted up to 2.2 mm height. The top transparent layer is then cast on top of the Soft Gel Layer. After this, the outermost layer dyed with black ink is inserted to cover the sides of the sensor from dust and wearing off. Finally, the PDMS polymer is cast on top of the transparent layer to anchor the elastomer to the glass. The black cover layer incorporates extruded features for gripping the soft sensing pad from the sides  ($\sim1.5$ mm). The fabricated sensor is shown in Figure~\ref{fig:assembly}b and c.
%
% \begin{figure}[h]
% \centering
% \includegraphics[width=0.744\columnwidth]{figures_NUSense/manufacturing.png}
% \caption{The silicone molding-based fabrication of NUSense: the outermost layer was cured first in a  mold. Afterwards, painting masks for yellow and red inks were placed during the curing time. Finally, the pattern was covered with a softer transparent silicone rubber. All molds are 3D-printed.}
% \label{fig: manufacturing}
% \end{figure}
% \begin{figure}[h]
% \label{fig:assembly}
% \centering
% \includegraphics[width=0.3\textwidth]{figures_NUSense/assemble sensor.png}
% \caption{NUSense assembly schematics }
% \end{figure}
% \begin{figure}[h]
% \label{fig:Schematics}
% \centering
% \includegraphics[width=0.3\textwidth]{figures_NUSense/Concept_of_sensor.png}
% \caption{Schematics. NUSense sensor inside.}
% \end{figure}

% The layer itself is made of a soft 1~mm thick transparent silicone  mixed separately with yellow and red pigments

% It is covered from the top by a harder silicone mixed with black pigments. The bottom side of the sensor is rigid so that the top layer, i.e. the colored pattern, can move relative to the bottom thanks to the soft silicone between them

% \input{content/properties}
% \input{content/data_processing}
\section{Methodology}
\label{sec:data_proc}
In this section, we obtain a quantitative estimation of the shear deformation (see Section~\ref{sec:prince}) of the artificial tactile sensing surface.  We begin by providing background on parametric construction of surfaces using B-splines, followed by a detailed algorithm for processing the tactile image from the fabricated NUSense sensor.
\subsection{B-Spline Surface}
\label{subsec:b_spline} 
The displacement of surface for a soft sensing layer as in Figure~\ref{fig:concept}a (see Displacement of surface) may be represented by
\begin{equation}
\mathbf{S}(u, v) = \sum_{i=0}^{n} \sum_{j=0}^{m} N_{i,p}(u) M_{j,q}(v) \mathbf{P}_{i,j}
\label{eq:bspline}  
\end{equation}
where $\mathbf{S}(u, v)$ is a B-spline surface; $\mathbf{P}_{i,j}$ are the control points arranged in a grid; $N_{i,p}(u)$ and $M_{j,q}(v)$  are the B-spline basis functions of degree $p$ and $q$ in the $u, v \in [0,1]$ directions, respectively; $n, m$ are the number of control points in the $u, v$ directions, respectively. Please refer to ~\cite{gordon1974b} for additional details.

Construction of the surface involves a fine grid over $u, v$ to generate a smooth surface from the control points. These control points are obtained from the camera images, as described in next subsection.
\subsection{NUSense Data Processing}
\subsubsection{Preprocessing} The raw image from the sensor has a wide view captured through a fisheye lens camera as shown in Figure~\ref{fig:methodology_collage}A. We remove the fisheye distortion of the raw image by applying the camera matrix. This matrix is estimated from the parameters of the fisheye lens, such as the intrinsic matrix and distortion coefficients. The resultant image is the undistorted tactile image, $\Img$, which is later used for control point extraction (Figure ~\ref{fig:methodology_collage}B).
\subsubsection{Algorithms for Control Point Extraction} 
% The algorithm for control point extraction is outlined in Algorithm~\ref{alg:feature_extraction}.

The algorithm starts by cropping the region of interest from the original image, $\Img$, using predefined image coordinates and dimensions (x, y, width, height) . The resultant image is then converted to grayscale.

Next, a bilateral filter is applied to the grayscale image to reduce noise while preserving edges, which are detected using the Sobel filter. The combined Sobel image is thresholded to create a binary image. This binary image undergoes morphological operations, specifically dilation followed by erosion (closing), to fill small gaps and remove noise. These operations help create more solid shapes for contour detection. We apply contour detection~\cite{suzuki1985topological} and   extract the polygons of quadrilaterals as shown in Figure ~\ref{fig:methodology_collage}C. Then, we locate the corners of neighboring quadrilaterals and extract the midpoints between them, denoted $\mathbf{P}$, as shown in Figure ~\ref{fig:methodology_collage}D. We use these points to form the set of control points for the B-Spline surface defined in~\eqref{eq:bspline}.
\subsubsection{Sample from B-Spline Surface}
The resultant control points, $\mathbf{P}$, are fitted to the B-spline surface as described in subsection~\ref{subsec:b_spline} to construct $S(u,v)$. Then, we sample evaluated points from this surface: $\mathbf{s} \sim S(u,v)$, where $\mathbf{s} \in \mathcal{R}^{3 \times K}$ is a vector containing $K$ points.
\subsubsection{Metrics}
\begin{figure}[h]

\centering
\includegraphics[width=0.95\columnwidth]{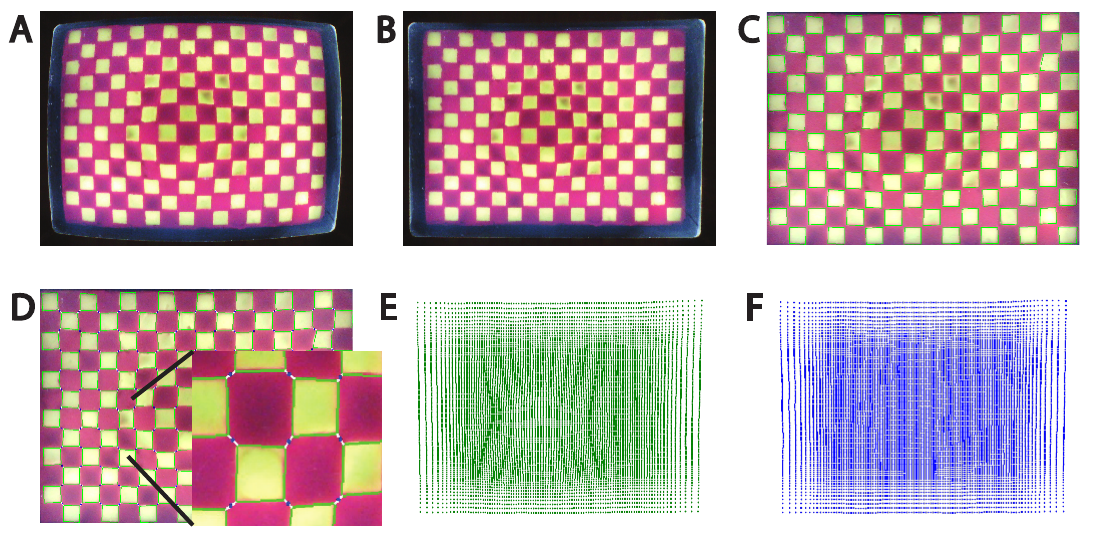}
\caption{Tactile image processing: A. Raw Tactile image from fish eye lens camera, B. Undistorted image, C. Quadrilaterals as a polygons around yellow markers, D. Midpoints extracted as control points for the B-Spline surface with magnified view, E. Sampled points from target B-Spline surface, F. Sampled points from reference B-Spline surface.}
\label{fig:methodology_collage}
\end{figure}

Here, we define the metric used to evaluate touch on the tactile sensor based on our observations of shear displacement~\eqref{eq:gamma} noted by the control points~\eqref{eq:bspline},
\begin{definition}[Shear Strain] 
Let $\mathbf{s}^{(\text{ref})}$ and $\mathbf{s}$ be the sampled surface points from the reference $S^{\text{ref}}$ and target $S$ B-Spline surfaces, respectively. Then the \emph{shear strain}  $\gamma_{\text{ss}}$ of the touch is defined as a scaled L2-norm:
\begin{equation}
    \gamma_{\text{ss}} = \alpha \sum_{i=1}^{K} ||s_i - s_i^{\text{ref}}||_2
    \label{eq:estim_gamma}
\end{equation}
where $\alpha$ is a constant scaling parameter.
\end{definition} 
For our setup, we set the degree for both the $u$ and $v$ directions to be $1$, i.e., $p=q=1$, and $\alpha = 18000^{-1}$. Additionally, we assume that the B-spline is projected onto the plane. The size of the control points corresponds to the number of quadrilaterals in the tactile image, which is $m=n=164$. Figure~\ref{fig:methodology_collage}E and Figure~\ref{fig:methodology_collage}F illustrate reconstructed surfaces with $s$ and $s^{\text{ref}}$, respectively. A visual representation of \eqref{eq:estim_gamma} applied to the surfaces is shown in Figure~\ref{fig:concept}c. The control point extraction from tactile images are implemented in Python using OpenCV library~\cite{bradski2000opencv}.

\section{Sensor Characterization}
\label{sec:characterization}
In this section we validate the proposed tactile imaging with a set of experiments conducted with a robot arm.

\subsection{Robotic Setup}
\begin{figure}[h!]
\centering
\includegraphics[width=0.85\columnwidth]{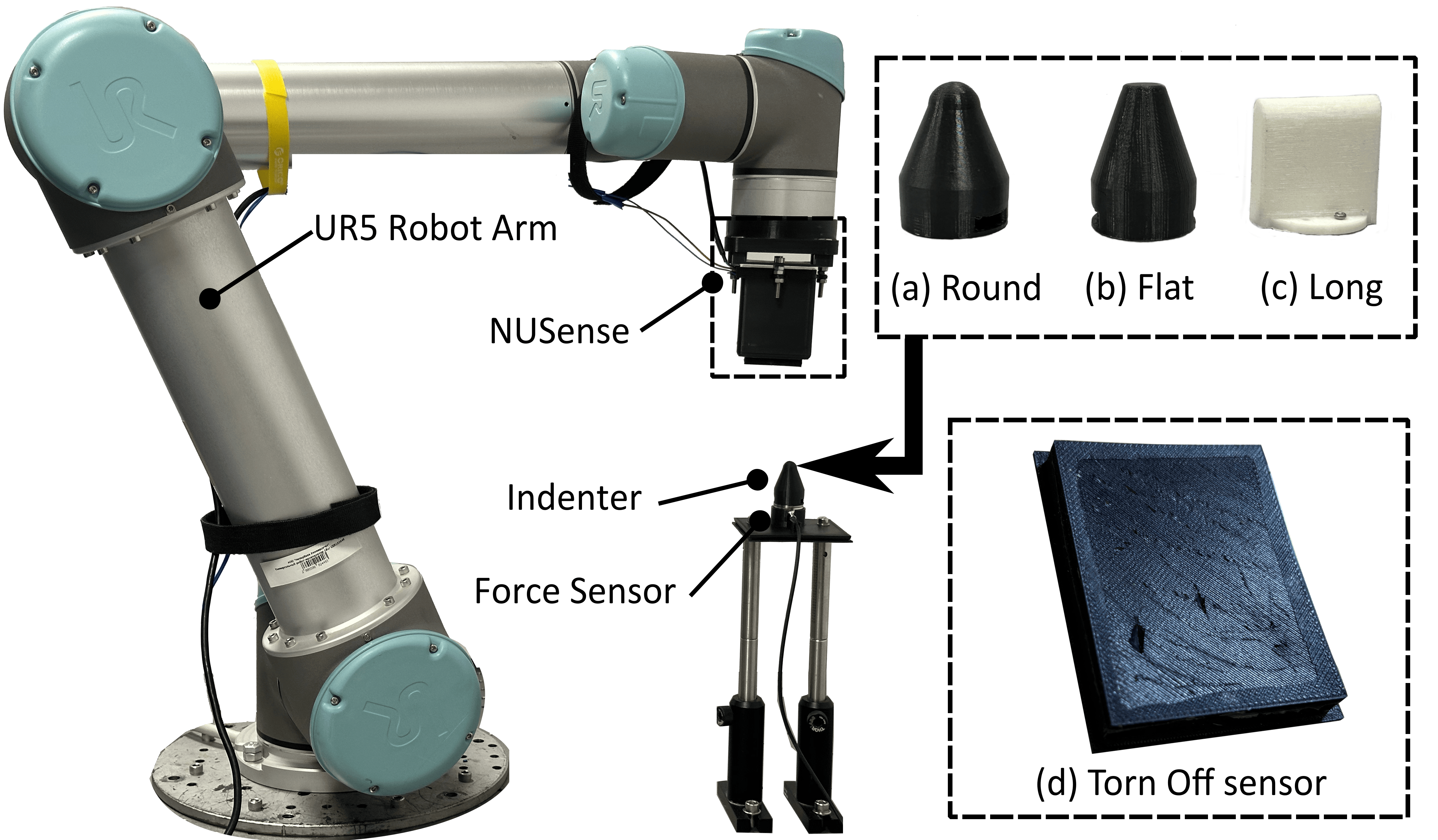}
\caption{\textbf{Left:} Experimental Setup. Indenters with three different areas of contact. Normal force response in the ranges 1 N to 3 N and 3 N to 8 N was obtained by (a) round and (b) flat tips, respectively. (c) Long tip was used for edge detection. (d) Torn off elastomer sensor.}
\label{fig:robotic_setup}
\end{figure}

We attached our NUSense tactile sensor onto a robot arm (Universal Robots; repeatability ±0.1 mm ). Edge-  and point-protruding indenters were secured on top of an optical breadbaord with mounting plate (Thorlabs). An industrial force/torque sensor (hex21, Wittenstein) was fixed between the plate and the indenters to calibrate the sensor output to SI units (Figure~\ref{fig:robotic_setup}). The robot arm was controlled using the position servoing mode at 125 Hz. The force/torque  sensor was sampled at the same rate to obtain ground truth force measurements. The communication interface with the workstation (Intel Xeon E5620 2.40 Ghz, 4 GB DDR3 memory, Ubuntu Linux) was implemented using Robot Operating System 1 (ROS). Camera images were sampled at 12 Hz. 

\begin{figure*}
\centering
\includegraphics[width=1\linewidth]{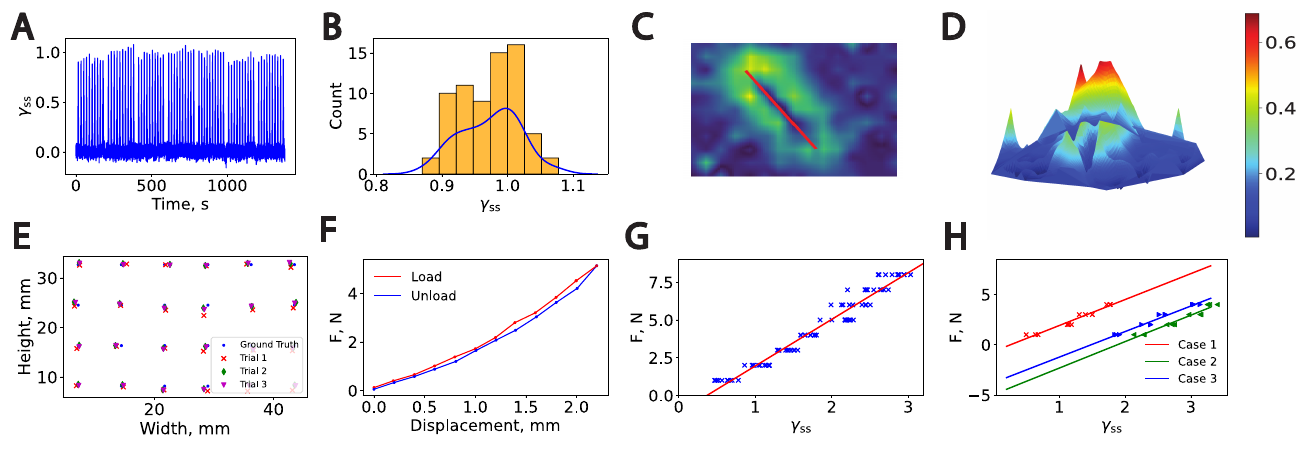}
\caption{\textbf{Experimental Results:}
\textbf{A}. Shear strain versus time
\textbf{B}. Histogram of the maximum deformations for each touch in the robustness experiment
\textbf{C}. Shear strain projection on 2D
\textbf{D}. Corresponding 3D visualization
\textbf{E}. Single indent touch localization
\textbf{F}. Hysteresis
\textbf{G}. Force calibration
\textbf{H}. Three cases for \textit{tore-off} sensor configuration.}
\label{fig:experiments_figure}
\end{figure*}

% Every indenter protruded the NUSense sensor at desired locations that were manually set by positioning them at the geometrical center and at maximum depth of protrusion corresponding to 8 N. 

Each indenter was positioned to protrude the NUSense sensor at desired locations, which were manually set at the geometric center. The maximum depth of protrusion corresponded to a force of 8 N.

\subsection{Experiments}
\subsubsection{Normal Force Estimation}
\label{sec:calibrate}

To assess NUSense's ability to measure the normal force, a round tip indenter (Figure~\ref{fig:robotic_setup}a) and flat tip indenter (Figure~\ref{fig:robotic_setup}b) with the tip size of 5 mm in radius compressed the soft pad up to 3 N and 8 N, respectively. Figure~\ref{fig:experiments_figure}G depicts the estimated shear strain \eqref{eq:estim_gamma} calibrated to the normal force for these intenders. The force correlates linearly with shear strain between $1$ and $8N$ , with a fitting function of $F = 3.09\gamma_{\text{ss}} - 1.14$. Lower values below $1N$ were not sensitive enough to produce deformation, while forces above $8N$ were avoided to prevent damage to the sensor.

\subsubsection{Single contact localization}
For both indenters (round and flat), the region of the shear deformation forms an annulus with the center at the point of contact. We evaluated the sensor's capability in localizing the contact. The indenter applied 6 N of pressure at 24 points on the sensor surface with the step of $6.4$ mm and $7$ mm in x- and y- axes. 

Figure~\ref{fig:experiments_figure}E shows the contact locations of three different trials on the sensor. Overall, touch was well-localized, albeit with a slight drift along the height of the sensor. The RMS error for contact localization in this experiment was $0.50 (0.09)$ mm.

\subsubsection{Hysteresis}
Due to the viscoelasticity of the silicone rubber, the reaction force during mechanical load and unload may differ. This behavior can be observed in Figure~\ref{fig:experiments_figure}F, which shows the hysteresis plot for NUSense. The robot arm moved at the constant speed of $0.5\frac{mm}{s}$ to protrude the intender with the round tip. The non-linearity in hysteresis increases with the softness of the silicone rubber. Thus, there is a trade-off between the sensitivity and bandwidth of the sensor, where the latter is important for force control.

\subsubsection{Robustness}
The sensor output should remain consistent during repetitive tasks. We set the robot to perform a cycling load. Figure~\ref{fig:experiments_figure}A shows the 70 touches on the NUSense and the corresponding maximum shear strain. The maximum shear strain exhibits relatively low variance, as shown in Figure ~\ref{fig:experiments_figure}B, indicating robust and repeatable sensor performance.

\subsubsection{Edge detection}
Edge indentation is another common type of contact. A long indenter was used to protrude the sensor with a force of 8 N at different orientations relative to the x-axis. 

The shear strain during contact exhibited a clear pattern (Figure~\ref{fig:experiments_figure}C). Principal Component Analysis (PCA) with two components was applied to the most deformed sampled points to extract the direction of the edge. The corresponding direction of the eigenvector for the largest eigenvalue indicates the angle of the indent and is represented as a red line. The red line is approximately 60 degree in clock-wise direction. A 3D visualization of the region of shear for this edge-contact is shown in 
 Figure~\ref{fig:experiments_figure}D. Note that the contact area is not deformed, while the shear strain is clearly visible around the contact.
 % Similar to~\cite{she2021cable},
\subsubsection{Force detection under sensor damage}
The outermost layer of the sensor may become damaged when grasping sharp objects. To test the robustness of the sensor in such scenarios, we intentionally cut this layer. The photo of the damaged sensor with the worn-off layer is shown in Figure~\ref{fig:robotic_setup}d. The soft part of the sensor can be replaced, and the sensor output should remain consistent across different soft pads.

We conducted an experiment with three cases: for Cases 1 and 2, we used different pads, while for Case 3 we used the intentionally damaged pad for the sensor. Figure ~\ref{fig:experiments_figure}H shows these three cases for 1, 2, 3, and 4 N contacts. Similar to the force calibration experiment, the observed shear strain is linearly correlated with normal forces, and all three cases exhibit a similar slope. The bias differences arise from various factors, such as manufacturing variations for Cases 1 and 2 and the torn-off sensor pad for Case 3. Overall, the sensor retains its ability to identify forces even in extreme cases where the sensor pad was damaged.

\section{Conclusion}
\label{sec:concl}

In this work, we presented the NUSense optical tactile sensor, designed for implicit force estimation through shear strain detection. Unlike traditional methods, our approach leverages the observation of shear strain via a vision-based system to infer the applied force. This novel sensing principle was validated through multiple experiments, demonstrating the sensor's ability to measure normal forces, localize contact points, and detect edge orientations with high repeatability and robustness. Our proposed method is subject to certain limitations. The soft pad, composed of different silicone rubbers, was considered isotropic, and the analysis assumed that the material remains within its elastic limits without punctures. Future work will focus on addressing these limitations, as well as exploring further improvements in material design and algorithmic approaches.

\bibliographystyle{./IEEEtran}
\bibliography{./IEEEabrv,./IEEEexample}
\end{document}